\documentclass[letterpaper,11pt,reqno]{amsart}

\theoremstyle{plain}

\theoremstyle{definition}

\theoremstyle{remark}

\usepackage{enumerate}
\usepackage{amssymb}
\usepackage[mathscr]{eucal}
\usepackage{graphicx,color}

\usepackage{hyperref}
\usepackage{dirtytalk}
\usepackage{xpatch}
\usepackage[export]{adjustbox}
\graphicspath{ {./plots/} }
\begin{document}

\author[
Nat Roth
Justin Wagle,
]{%
Nat Roth
Justin Wagle,
\\
\MakeLowercase{\{narot,justiwag\}@microsoft.com
}}

\title[One of these (Few) Things is Not Like the Others]{One of these (Few) Things is Not Like the Others\\
\tiny{Few-Shot Image Learning and Outlier Detection}}
\date{\today}

\maketitle

\newcommand{\Loss}{\mathcal{L}}

\begin{abstract}

To perform well, most deep learning based image classification systems require large amounts of data and computing resources. These constraints make it difficult to quickly personalize to individual users or train models outside of fairly powerful machines. To deal with these problems, there has been a large body of research into teaching machines to learn to classify images based on only a handful of training examples, a field known as few-shot learning. Few-shot learning research traditionally makes the simplifying assumption that all images belong to one of a fixed number of previously seen groups. However, many image datasets, such as a camera roll on a phone, will be noisy and contain images that may not be relevant or fit into any clear group. We propose a model which can both classify new images based on a small number of examples and recognize images which do not belong to any previously seen group. We adapt previous few-shot learning work to include a simple mechanism for learning a cutoff that determines whether an image should be excluded or classified. We examine how well our method performs in a realistic setting, benchmarking the approach on a noisy and ambiguous dataset of images. We evaluate performance over a spectrum of model architectures, including setups small enough to be run on low powered devices, such as mobile phones or web browsers. We find that this task of excluding irrelevant images poses significant extra difficulty beyond that of the traditional few-shot task. We decompose the sources of this error, and suggest future improvements that might alleviate this difficulty.

\end{abstract}

\section{Introduction}

As machine learning becomes a growing part of companies' products and users' experiences, it is increasingly important to build models which can personalize to individual users' preferences and judgment. Depending on a user's profession and personality, the same piece of content can have very different meanings. A picture of a hotel room could be a fun vacation memory for most, but for the hotel staff it might be a reminder that a room needs cleaning or repair. Further, if we were to try to train a model across a large group of users to recognize whether a picture is work related, there would likely be very few boutique hotel workers labeling pictures of rooms. A naive algorithm would likely conclude that pictures of a small bed and breakfast are not business related, and would be right in that assumption, almost all the time. Still, this approach would fail the small business owner or their employees who want some type of AI to help organize their work pictures. 

To solve this problem, there has been an explosion of cloud offerings which allow users to train custom models based on the images they provide \cite{azurecustom},\cite{sfcustom},\cite{googlecustom},\cite{awscustom}. However, at a time when a majority of the American public does not trust tech companies' use of AI \cite{zhang2019artificial}, these approaches all demand that users send their data off premise to some company's server. Depending on legal restrictions, whether the cloud service is run by a rival company, and employer policies, individual employees may not even be allowed to aggregate work related images on many of these services. 

\begin{figure}[ht]
\includegraphics{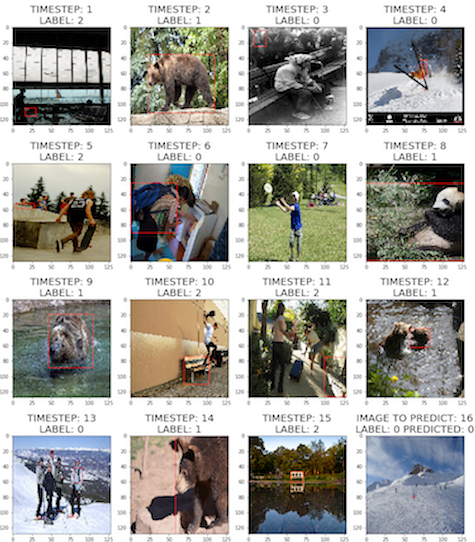}
\caption{Example episode with classes of bear, bench, and backpack. Bounding boxes are not supplied to model, but shown to help visualize objects of relevance}\label{fig:fE1_2d}
\end{figure}

This friction and lack of trust opens the door for algorithms that can learn personal preferences, do so on device, and allow the user to update their model quickly. One solution to this problem comes from the field of few-shot learning. The standard few-shot learning tasks in the literature assume that a few example images for a couple categories are presented, and then a single image is presented to be labeled, as can be seen in \ref{fig:fE1_2d}. This setup is often referred to as a training episode. 

We extend this standard setup to allow the query image to potentially not belong to any of the previously presented classes. This change is motivated by the observation that many user facing features require an ability to filter out irrelevant images. Returning to the case of a hotel worker, their phone may be a mix of work images they want categorized, such as pictures of beds, hotel rooms, receipts, as well as personal images they do not want classified, such as \say{selfies}, screen-shots they shared with friends, and family pictures. The ideal scenario would be that the user could provide a small number of labels for the categories they did care about, and the model would learn both to classify those as well as filtering out the images that are \say{junk} or do not belong to any of those categories.

To accommodate this extra difficulty, we propose a model that simultaneously learns to categorize images in an episode as well as predicting whether the image should be classified at all, or is rather \say{junk}. Our model first encodes images and then averages all images with the same category, as in the work on \say{prototypical networks} \cite{snell2017prototypical}. New images are then assigned to the class their encoding is closest to. In addition, the distance between an image's encoding and these category averages is fed through a linear layer in order to determine whether the image is actually \say{junk}. To our knowledge, this adaptation of the standard few shot task to include \say{junk} images, as well as the addition of this final linear layer to the \say{prototypical networks} work are novel. To more closely mimic a real world experience, we run this model on the MSCOCO image dataset \cite{lin2014microsoft}, a dataset of images known for being messier than the standard few-shot image datasets. We observe lower accuracy on predicting whether an image is \say{junk} than we see on the standard prediction task and discuss why this might be. We analyze the sources of error in our model and propose future extensions to address some of the difficulties this new task presents. 

Ultimately, our contributions are as follows: we present a novel task and adapt existing work to tackle it. We provide a concrete example of how few-shot performance can change when moved to a more difficult dataset. We also report results on how few-shot and \say{junk} performance changes as model size is reduced in order to fit on device, which can guide practitioners evaluating whether to move server-side models to the phone or browser.

\section{Related Work}

\subsection{Few Shot Learning and Fast Weights}

There is a long history of learning quickly from small amounts of data, a problem often referred to as few-shot learning or the learning of \say{fast weights} \cite{schmidhuber1993ratiolearning}, \cite{hinton1987using},\cite{munkhdalai2018metalearning}, \cite{ba2016using},\cite{miconi2018differentiable}, \cite{kang2018few}. 

Our approach most closely mirrors that of \cite{snell2017prototypical}, which proposes averaging the encodings for all examples that belong to a specific class and refers to that average as the \say{prototype} for that class. Predictions for new classes are made by finding the nearest \say{prototype}. This approach is related to \cite{vinyals2016matching} which looks over all points, rather than averaging by class first. In both these setups, the task requires learning a good embedding function, in which nearby points or the average of points, are predictive for future examples.

\subsection{Few Shot Detection of Outliers}

The problem of determining whether a data point is an outlier or anomaly is also well-studied one \cite{scholkopf2000support},\cite{chandola2009anomaly}, \cite{mahadevan2010anomaly}. There has been significant research regarding using neural networks to determine whether an image is novel or belongs to some unseen class \cite{socher2013zero}, \cite{fu2018recent}, \cite{zmaurernovelty}, \cite{sabokrou2018adversarially}. However, this work often assumes that there is a medium to large number of non-anomalous images. For example, Socher et al. \cite{socher2013zero} learn a manifold for images in their training classes, which number in the thousands, and then fit a model to determine if a image lies on that manifold or not. In \cite{zmaurernovelty} the author clusters a large number of embedded training images and uses a variety of techniques to learn what is an acceptable distances between points and cluster centers. Like ours, this work focuses on approaches that could run on device, but its approach assumes that the cluster centers are learned off-line from many examples at train time and are then frozen and sent to device for inference. 

Our problem, while quite similar, differs in that we assume that there are only a very limited number of examples that the model can use at inference time to estimate the center of a class and the manifold related images might lie on. While previous work may do fairly well in this limited training data regime, it is, to our knowledge, not the focus of any of that work or directly reported on. 

Previous few-shot learning work could be extended to deal with outlier detection. \cite{vinyals2016matching} could be updated to randomly sample some novel examples. Alternatively, a threshold could be applied to determine how far away a point must be from its neighbors to be classified as an outlier. In our approach, we extend the work in \cite{snell2017prototypical} and train a model end to end to learn to classify images based on few examples, while simultaneously learning a cutoff for whether an image is \say{junk} based on its distances from \say{prototype} centers. We describe this approach in more detail below. 

\subsection{On Device Computing}
There are many benefits to being able to run such a model on a user's device. Users are growing increasingly skeptical of technology companies' use of their data. Users may be more willing to label personal images if that data is not aggregated remotely. Running models on device also saves server resources and allows models to effortlessly scale as the user base grows. 
While running image classifiers is traditionally a fairly computationally intensive task, there is a quickly growing ecosystem of tools and body of research that makes it easier to run models on device. We leverage the work on MobileNet \cite{howard2017mobilenets}, \cite{sandler2018mobilenetv2}, which introduces a smaller and faster convolutional network that still performs well on standard image recognition tasks. We employed the smallest Tensorflow provided pre-trained model with the MobileNet version 2 architecture \cite{tensorflowmobilenetv2}, as the other variants ran too slowly on the lowest end devices we tested.

\section{Experiment Setup}

\begin{figure}[ht]
\centering
\includegraphics{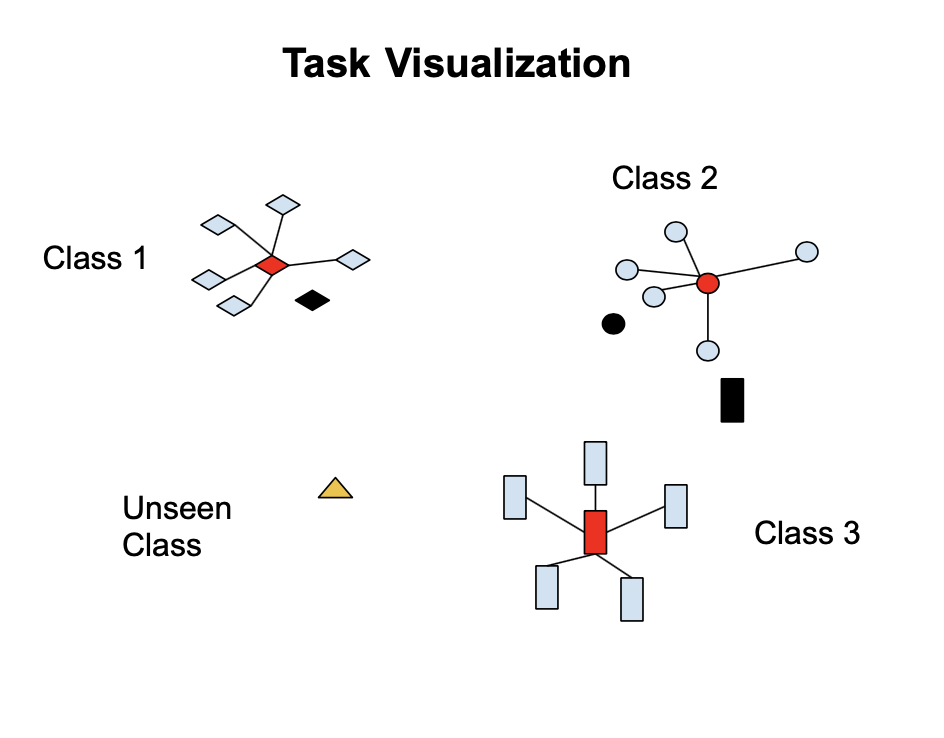}
\caption{Visualization of five shot task. Diamonds, circles, and squares represent observed classes, with observed examples in light blue and estimated class means in red. The triangle comes from a class not seen at train time, and should be classified as junk. The examples in black represent test time images and will be classified as belonging to the nearest class center or junk, if they are far away from all centers. The black square is an outlier and will be misclassified as it is closer to the circle class center than the square class center. The black circle and diamonds will be correctly classified.}\label{fig:f_viz}
\end{figure}

\subsection{Episode Construction}
We use the 2017 MSCOCO image dataset and the associated labels as our source of ground truth. MSCOCO provides a set of images along with labels denoting what objects are in each image, as well as train, validation, and test splits for those images. We choose MSCOCO rather than using datasets from previous few-shot learning work, like miniImageNet \cite{vinyals2016matching} as the images in MSCOCO were purposely created to be more cluttered and contain multiple objects. We believe this multiplicity and ambiguity more closely matches our datasets of interest, such as a user's camera roll. 

We set up our training episodes in a similar way to \cite{vinyals2016matching}; in particular, of the 80 MSCOCO objects, we reserve 15 for testing, 8 for validation, and 57 for training. Our training and validation images are taken from the train split for MSCOCO, while the images in our test set are drawn from the provided MSCOCO validation split of the images. Note that we use the provided validation split for our test set as the labels for the MSCOCO provided test split are withheld. 

Figure \ref{fig:fE1_2d} illustrates the procedure for creating a training episode. We first sample 3 object classes at random from the train object classes. We then present the model with a small number of images (\say{shots}) from each class. The model then embeds these images and constructs a representation vector for each class, as we will describe below. The model is then shown a single test image. With 75$\%$ probability, this image is drawn from one of the three training classes; with 25$\%$ probability, it is drawn from a random other class that was not seen in the episode (\say{junk}). The model must then predict one of four labels: one of the three classes, or \say{junk}. We train the model using back-propagation with a cross-entropy loss. 

At test and validation time, our setup is identical, except that rather than choosing from images that contain the 57 train objects, we choose from images that contain the 15 or 8 images in our test and validation sets. This allows us to see the extent to which the model is able to generalize and group together objects it was not directly trained on.

\subsection{Overlapping Classes}

Since all classes may not be equivalently easy to classify or may not equally help the model generalize to new classes, we perform 12 random splits of the classes, with a different, random subset being withheld for validation and testing each time. We then average results across all splits. MSCOCO images consistently have multiple labeled objects in them, so we must determine how to train on images that include objects in both our train and test sets. We don't want to only look at images only containing a single class, as there are relatively few of those in MSCOCO and since we chose the dataset specifically because there are multiple objects in most images.

Instead, we follow the following procedure. First, we use the provided training dataset split from the MSCOCO to create our train and validation sets. For every image in the provided MSCOCO training dataset split, if the image has only objects in our train or validation set, it is automatically assigned to that set. If the image has training objects, but also includes objects in the validation or test sets, we flip a coin and with 50 percent probability, we keep it as a training image. Otherwise, if the image has an object in the validation set, we add it to our validation data, and ignore the image if it has no objects in the validation set. 

For every image in the provided MSCOCO validation dataset split, if it contains any test object, we include it in our test set. If the image has no objects in the test set, we ignore it. 

This procedure means that there may be training images that contain objects in our test set. For example, consider a case where \say{human} is a training object, and \say{tennis racket} and \say{fork} are test objects. It is possible our train set contains some humans holding tennis rackets and forks, as well as engaging in other activities. This approach is most similar to the random splits of ImageNet \cite{russakovsky2015imagenet} used in \cite{vinyals2016matching}, in which 118 objects were randomly removed from ImageNet. Though ImageNet is generally a purer dataset than MSCOCO, there are still pictures with multiple objects. For example, many pictures of tennis balls or soccer balls contain dogs playing with them. So, even if not trained directly on dogs, a network trained on tennis balls may still see many dogs while running on ImageNet. 

\section{Model}

\subsection{Architecture}

We closely follow the work in \cite{snell2017prototypical}, but force our model to predict whether an image doesn't belong to any of the provided classes, a categorization we refer to as \say{junk}. We present a visualization of the model and task in Figure \ref{fig:f_viz}

We first embed each image using the final layer of a convolutional neural network, which we refer to as $F$ below. We experiment with both pretrained networks and networks that we train from scratch, as we describe in Section 4.2. For each of our $K$ classes, we loop over all examples in that class, a set we refer to as $C_k$, and project these embeddings down to a 128 dimensional space using $g$. We average these projection within their provided class labels, resulting in a single class vector we define as $p_k$ in Equation \ref{eq:center}. 

\begin{equation}\label{eq:center}
p_k = \frac{1}{|C_k|} \sum_{x \in C_k} g^\intercal F(x) , {k \in 0... K} 
\end{equation}

To classify a query image, $x'$, we again embed the image using F. We then calculate an L2 distance between each of the class vectors and the embedded image vector, as in Equation \ref{eq:distance}.

\begin{equation}\label{eq:distance}
S_{C_k}(x') =  - \lVert g^\intercal F(x') - p_k \rVert, {k \in 0... K}
\end{equation}

We feed these distances through a linear layer to get a score for whether the image does not belong to any of the original classes. We learn two scalars, $b_{distance}$ and $b_{magnitude}$, as in Equation \ref{eq:junk}.

\begin{equation}\label{eq:junk}
JUNK(x') = b_{distance} \sum_{k=1}^{K} \lVert g^\intercal F(x') - p_k \rVert  + b_{magnitude} \lVert g^\intercal F(x') \rVert
\end{equation}

We note that $b_{distance}$ upweights our junk score by some multiple of the total distances between the query image and our class vectors. We expect this to be a positive number as the farther an image is from the class centers, the more likely it is to be \say{junk}. 

On the other hand, $b_{magnitude}$ updates our \say{junk} score by some multiple of the magnitude of the query image embedding. This allows the model to control for the fact that distances from class centers may be larger if the vector itself is of a larger magnitude. If we multiply all our embeddings by some arbitrary scalar, we would also multiply our distances by that scalar, increasing our \say{junk} score. However, by adding $b_{magnitude}$, the model can learn to counteract that arbitrary scaling, as the contribution of $ b_{magnitude} \lVert g^\intercal F(x') \rVert $ would also increase. The $b_{magnitude}$ term should be negative to balance out the positive contributions from $b_{distance}$.

The resulting score and the per-class distances are fed through a softmax function to yield the final probabilities, as seen in Equations \ref{eq:junkp}, and \ref{eq:nonjunkp}. 

\begin{equation}\label{eq:junkp}
P_{JUNK}(x') = \frac{e^{JUNK(x')}}{e^{JUNK(x')} + \sum_{k=1}^{K}{e^{S_{C_k}(x')}}}
\end{equation}

\begin{equation}\label{eq:nonjunkp}
P_{C_k}(x') = \frac{e^{S_{C_k}(x')}}{e^{JUNK(x')} + \sum_{k=1}^{K}{e^{S_{C_k}(x')}}}, {k \in 0... K}
\end{equation}
\\

In practice, we report results for $K=3$. This model requires a constant amount of memory and compute time in order to classify a new point. New examples for a class only require updating a single center and so can also be incorporated in constant time. This is allows the model to scale easily on the edge as more user behavior is observed.

\subsection{Training}

Training is done using the ADAM optimizer \cite{kingma6980method}. We begin with a learning rate of .0005, which is exponentially decayed over the course of training. Episodes are grouped into mini-batches of 8 and the model is trained for a total of 32000 mini-batches, with the final model selected using early stopping based on validation loss. In the case of using a pre-trained model, such as mobile net or VGG-16 \cite{simonyan2014very}, we freeze the CNN layers and train only a simple linear projection, as well as the linear layer mapping the L2 distances to the \say{junk} score. In the case of training a network from scratch, training is done end to end.

\subsection{Pre-Trained Network}
We note that while our image embedding networks (MobileNet, and VGG-16) were not trained on MSCOCO directly, they were trained on the ILSVRC-2012 \cite{ILSVRC15} dataset, which has many of the same types of objects. While there are substantial differences between the datasets, as discussed above, this does partially break from the normal few-shot learning setup.

This experiment still has real-world relevance for several reasons. At train time, the model does not need to pick up on objects in the test class, and in fact may sometimes want to ignore them since they do not directly contribute to the train task. The projection from the image features to our task specific embedding may wash out the objects in the test set while training. It is unclear how well the model should be able to do on tasks that require recognizing these objects at test time. 

The model must also learn to project these feature embeddings down in a way which is robust to errors in the original network encodings and which perform well when aggregated via averaging. If a single image is an outlier for a class, or the model misidentifies the objects in that image, the resulting representation ought minimally poison the class prototype. Especially as the pretrained model is moved away from its original dataset to a more complicated one, this robustness is important. 

Finally, it is useful to have some benchmarks that allow us to determine how much of our error is a result of a poor understanding of images rather than other factors. If we only look at a network trained from scratch, much of our error would be the result of the failings of that encoder, rather than our model architecture, the inherent ambiguity of our images, or the training setup itself.

\section{Results}

\subsection{Tasks}
We report performance on a variety of tasks. In Tables \ref{tab:fE1_2a} and \ref{tab:fE1_2b} we report accuracy on both \say{junk} and \say{non-junk} prediction for both the 1 and 5 shot tasks across various image embedding architectures. In section $5.4$, we compare our performance on MSCOCO to previous work. Finally, in Table \ref{fig:fE1_2c} we focus on accuracy purely as a function of \say{shots} and compare to simulated data.

\subsection{Junk Task}

We analyze our performance on simultaneously filtering out \say{junk} images and classifying images of user interest. We compare performance of various models on both 1 and 5 shot classification tasks in order to get a sense of how much increasing the number of examples increases performance across all model architectures. In all cases, we have 3 classes we present to the model and one \say{junk} class, resulting in a 4 way classification problem. We report \say{non-junk} accuracy, which is the accuracy we observe on episodes where \say{junk} is not the correct label. We also note \say{junk} accuracy, which is the probability that our model says an image is \say{junk} given it is actually \say{junk}. Finally, we report the AUC of the models in determining whether something is \say{junk} or not. Note that in all experiments, 25$\%$ of images at train time are junk, and randomly guessing would yield 25$\%$ accuracy on both \say{junk} and \say{non-junk} images and an AUC of $.5$ . All experiments are run on the MSCOCO dataset.

\begin{table}[ht]
\begin{tabular}{|p{2 cm}|p{3.1cm}|p{3.1cm}|p{2.1cm}|}
\hline
Model & Percent Non-Junk Accuracy 1-Shot & Percent Junk \newline Accuracy 1-Shot & AUC 1-Shot \\
\hline
VGG&62.7&30.3&.691\\
\hline
MobileNet&57.7&24.4&.659\\
\hline
CNN&47.9&14.8&.590\\
\hline
\end{tabular}
\caption{Results for 1 shot task.}\label{tab:fE1_2a}
\end{table}

\begin{table}[ht]
\begin{tabular}{|p{2 cm}|p{3.1cm}|p{3.1cm}|p{2.1cm}|}
\hline
Model & Percent Non-Junk Accuracy 5-Shot & Percent Junk \newline Accuracy 5-Shot & AUC 5-Shot \\
\hline
VGG&73.3&39.1&.766\\
\hline
MobileNet&66.9&37.2&.735\\
\hline
CNN&56.4&23.5&.634\\
\hline
\end{tabular}
\caption{Results for 5 shot task.}\label{tab:fE1_2b}
\end{table}

\subsection{Junk Performance}
Performance on the \say{junk} class is significantly worse than accuracy on the standard task, consistently trailing by around 30$\%$. This makes sense, as classifying an image in the \say{junk} task is harder than adding another class to the problem. In the case of choosing between only three classes, the model must just make sure that an image is closer to its own class, in expectation, than any random three other classes. In the case of \say{junk} it must not only do that, but determine what counts as far enough away, a cutoff that may vary between classes. With the inclusion of \say{junk}, if a point is now far away from all 3 classes, but clearly closer to one of them than the others, it is no longer trivially easy to classify it as belonging to the nearest class. Instead, the model must pick between the nearby class and \say{junk}.

Our current setup has no direct way of knowing the variance of images within a class. Instead it has to use the same learned parameters, $b_{distance}$, for the distances for all classes. An image may be close to a class center in an absolute sense, but if that class has very small variance, it may still be the case that the image does not belong to that class and is instead \say{junk}. 

Future work might try to feed in the average distances of images in a class from their cluster centers, which would allow the model to fit to differences in variances between groups of images; however, such an approach might be noisy or unworkable when very few images are present and is useless in the case of having only a single image. Alternatively, the model could attempt to learn some global parameters to determine that certain types of images belong to more varying classes. For example, maybe pictures of food show up in a narrow range of situations, as they are always pictured up close and on plates. On the other hand, pictures of animals may be more varying as they can be seen with humans, or in the snow, grass, and water. The model could then guess if a new object seems to have more in common with high variance objects, or lower variance objects and score the \say{junk} class accordingly. 

\subsection{Comparison to Standard Task}

To get a sense of the relative difficulty of our dataset, we remove the \say{junk} class and compare our model performance on MSCOCO to that of the original \say{prototypical networks} paper on miniImageNet. We compare on a 5-way classification problem and observe that our accuracy is comparable but moderately worse. When using MobileNet as our embedding architecture, we see similar 1 shot accuracy, as our model achieves 48$\%$ accuracy on our MSCOCO dataset, while the previous work had 49$\%$ accuracy on the miniImageNet dataset. This gap widens as we see 60$\%$ accuracy when using 5 shots instead of the 68$\%$ the original work reported. We believe much of this gap is due to the increased difficulty and ambiguity of the MSCOCO dataset which has been noted in previous work. In particular, the original MSCOCO paper \cite{lin2014microsoft} points out that on an object detection task performance drops by half when moving from the PASCAL VOC dataset \cite{everingham2010pascal} to MSCOCO. Recent few shot work on objection detection in \cite{kang2018few} also notes that MSCOCO presents a high degree of difficulty for few-shot learning. Due to this increased difficulty, we believe performance on MSCOCO yields a more realistic sense of how a few-shot learning model might perform on real user data, and is a useful point of comparison for anyone seeking to deploy such a system. 

\section{Analysis $\&$ Discussion}

We can decompose our error on this task into two parts: the variance between images which ought be in the same group and our uncertainty in the actual correct center of that group. We may view each class of images as being represented by some high dimensional Gaussian, a piece of intuition \cite{snell2017prototypical} discusses. Under this assumption, a new image may be far away from its estimated class center if either the class itself is high variance or if we have not observed enough other images in the class to reliably estimate the center, as may often be the case in our few-shot experiments. In the simplest case of a one-dimensional Gaussian where we have perfect information about a class, a new point's expected squared distance to its class center  is the variance of that class. On the other hand, in the case where we must estimate the class center on a single example, as in one shot learning, our new point's expected squared distance to the center is two times the variance of the class. This points to the intuition that performance can be significantly improved by either more accurately calculating the mean of a class while fixing the variance or by simply reducing the variance within a class. We examine both below.

\subsection{Trends in Model Performance}

\begin{figure}[ht]
\centering
\includegraphics[center]{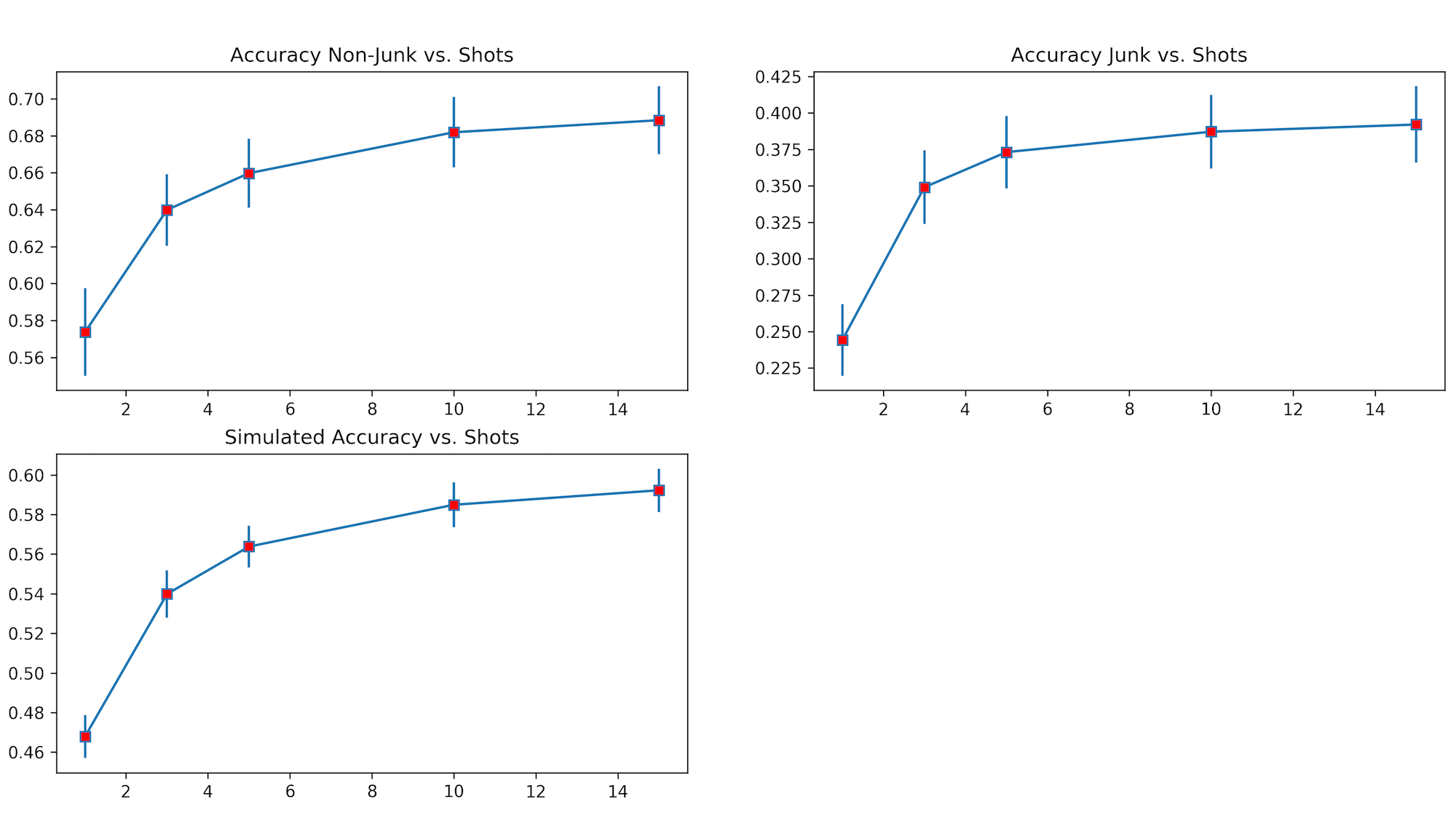}
\caption{Change in accuracy as a function of shots, error-bars represent 95$\%$ confidence interval}\label{fig:fE1_2c}
\end{figure}

We first measure the extent to which getting more accurate estimates of class centers aids performance. In particular, we calculate the performance of the MobileNet model as a function of the number of shots used for training. As we see in Figure \ref{fig:fE1_2c}, the added value of having the user provide more examples for a class is quite high for the first 5 examples or so, but tails off quickly after that. At a high level, this trend matches our intuition, namely that the improvement in performance should be some concave, rather than linear function of shots. If we envision our \say{prototypes} as being drawn from some high dimensional Gaussians, then we would expect the error in our estimation of the means of those Gaussians to roughly go down with the square root of the number of example.

To verify this, we simulate our same few-shot learning task on data drawn from a set of Gaussians. We randomly generate means for 15 Gaussians, and choose 3 at a time. For each Gaussian, we randomly sample K points, where K is the corresponding number of shots. We then randomly choose one of those K Gaussians, draw our query point from it, and assign that point to the closest class center. We see that our curve, pictured in the bottom left of Figure \ref{fig:fE1_2c} closely mirrors that of our actual performance. In the case of our simulation, our accuracy has peaked and adding significantly more points will not improve it much. This results because there is some overlap between the Gaussians, so some points may be sampled from one Gaussian, but actually be closer to the center of another. Such error is unavoidable without decreasing the inherent variance of the Gaussians themselves. This serves as a useful rule of thumb for practitioners deploying these models. If accuracy stops increasing as more examples are shown to a model, it is likely time to start working on improving the image embedder or otherwise reducing the noise in the images you are grouping together.

From a user experience perspective, this quick learning is a mixed blessing. It is good that the model learns quickly as users may be more willing to provide small number of examples, rather than tediously labeling dozens or hundreds of images, especially if those initial examples lead to rapid and visible improvement in the model. On the other hand, if performance plateaus, some users may become frustrated. This frustration may be amplified if they have elevated expectation based on how quickly the model learned to start. It is also possible that users may respond to the limitations of the model and make their groupings of images more pure.  

\subsection{Performance and Embedding}

While it is straightforward to measure the effect of adding more examples per class and thus more accurately estimating the centers of a class, it is less easy to measure the actual variance of the classes. We hypothesize one major source of variance comes from failures of our image recognition model. Two images may look similar to a human, but if the model does not embed them near to each other, our classes will be more spread out in the learned feature space. Under a very low quality model, all images would be randomly distributed in the space, and classes would have very high variance. A higher quality model will embed similar pictures nearby and thus likely lead to classes with lower variance, assuming classes are truly groups of similar images. Using embedder quality as a proxy for variance, we can more rigorously look at the effect of decreasing variance within the classes of images. 

Unsurprisingly, we see from both Table \ref{tab:fE1_2a} and Table \ref{tab:fE1_2b} that the larger VGG architecture significantly outperforms the other models. VGG is too large a model to be realistically deployed on many edge devices, but provides a baseline of how well a very powerful image recognizer could perform. We see that the much smaller MobileNet model performs worse than VGG, around .03 AUC lower and 5$\%$ less accurate. VGG learns far more quickly, as it achieves similar or even slightly better accuracy when show 5 images than MobileNet achieves when shown 15 images. 

This is a smaller discrepancy than we see between the CNN model trained from scratch and MobileNet, which is around .1 AUC and 10$\%$ less accurate. The CNN model is trained on just our training set in MSCOCO, which contains far fewer and noisier images than the ILSVRC dataset, on which VGG and MobileNet were trained. The size of the training data and the fact that CNN is not pre-trained to recognize objects similar to those in the test set likely explains much of the difference between it and MobileNet. 

The gap between the models implies that a significant bottleneck in performance on this task is the image representation quality. When the number of examples per class is held constant, we can still increase accuracy by over 15$\%$ by improving our image embedder. This jump is fairly close to the roughly 12$\%$ improvement we see when increasing the number of examples per class from $1$ to $15$. Again, we hope these numbers can help guide those trying to deploy these models, and give them a rough sense of the trade offs of using smaller models. Depending on your user experience, privacy constraints, and server costs, it may be worth it to run a smaller model if it only requires asking the user to label $10$ more images to achieve similar performance.

\section{Conclusion}
As the field of few-shot learning moves forward and into the hands of real users, fleshing out approaches to the problem of \say{junk} classification will become increasingly crucial. We see that our performance is greatly effected both by the quality of our image embeddings as well as the number of examples we have seen. In the case of the latter, we observe a concave trend to our improvement as a function of shots that is consistent with the hypothesis that our image embeddings are represented by some high dimensional Gaussians. We conclude that there are diminishing returns to observing more shots and that future improvements must come from improving our image encoders, which we hypothesize has the effect of tamping down the within class variance. We also believe that future work should explore new mechanisms for modeling the differences between the variances of different and that doing so could help significantly on the challenging task of \say{junk} prediction. We hope that our paper can help encourage further work into this  and similar problems. 

\bibliographystyle{amsplain}
\bibliography{ml}

\end{document}